\documentclass[conference]{IEEEtran}

\usepackage{cite}
\usepackage{amsmath,amsfonts,amssymb}
\usepackage{graphicx}
\usepackage[caption=false,font=footnotesize]{subfig}
\usepackage{booktabs}
\usepackage{siunitx}
\usepackage{hyperref}

\title{Structural Preservation Governs Data Augmentation in Deep Learning-Based Laser Speckle Material Classification}

\author{\IEEEauthorblockN{Mohamed Abdallah Salem}
\IEEEauthorblockA{\textit{College of Engineering} \\
\textit{North Dakota State University}\\
Fargo, ND, USA, 58102 \\
ORCID: \href{https://orcid.org/0009-0004-4673-420X}{0009-0004-4673-420X}}
\and
\IEEEauthorblockN{Nourhan Zein Diab}
\IEEEauthorblockA{\textit{College of Computer Science and Engineering} \\
\textit{New Mansoura University}\\
New Masoura, Egypt \\
nourdiab736@gmail.com}
}

\begin{document}

% \raggedbottom

%==========================  
\IEEEoverridecommandlockouts
\IEEEpubid{\makebox[\columnwidth]{ 979-8-3315-8488-7/26/\$31.00~\copyright~2026 IEEE \hfill}
\hspace{\columnsep}\makebox[\columnwidth]{ }}
\maketitle
\IEEEpubidadjcol
%===========================

\begin{abstract}
Data augmentation is routinely used to improve generalization in image classification, but the assumptions underlying standard policies are poorly matched to coherent imaging. Laser speckle patterns are not generic textures; they arise from coherent interference, and their discriminative content is carried by structured stochastic spatial and frequency statistics. This study examines how controlled augmentation perturbations influence speckle-based material classification on the SensiCut dataset. We train ResNet18 and EfficientNet-B0 under a parametric augmentation framework comprising rotation, Gaussian blur, independent Gaussian noise, spatially correlated speckle-aware noise, intensity jitter, and spatial masking, and evaluate test performance using macro F1-score averaged over three random seeds. Separate ordinary least squares models link augmentation parameters to performance for each architecture. Across both models, Gaussian blur exerts a strong negative effect ($p<0.001$), indicating that low-pass filtering suppresses high-frequency structure that is informative for material discrimination. Independent pixel-wise noise is likewise harmful ($p=0.003$ for EfficientNet-B0 and $p=0.001$ for ResNet18), consistent with disruption of local spatial coherence. In contrast, spatially correlated perturbations yield significant positive coefficients ($p=0.004$ for EfficientNet-B0 and $p=0.001$ for ResNet18), showing that variability can improve robustness when it preserves speckle organization. The fitted models explain a substantial fraction of performance variation ($R^2=0.796$ for EfficientNet-B0 and $R^2=0.879$ for ResNet18). These results show that, in laser speckle imaging, augmentation effectiveness is determined primarily by structural preservation rather than perturbation magnitude. The findings motivate physics-aware augmentation design for coherent optical sensing.
\end{abstract}


\section{Introduction}

\IEEEPARstart{L}{aser} speckle sensing exploits the fact that coherent illumination interacting with a rough or heterogeneous surface produces an interference pattern whose spatial statistics depend on the underlying scattering structure. In material recognition settings, those patterns are informative precisely because they retain sensitivity to microstructure, surface finish, and optical response rather than to conventional object appearance alone. Prior work has shown that speckle images can support material-aware sensing and classification, including in the SensiCut system for laser cutting, but the signal is fundamentally different from the semantic content usually assumed in natural-image classification pipelines \cite{boas2010laser,briers2013laser,dogan2021sensicut}. In speckle imagery, class-relevant information is embedded in local coherence, interference-driven texture, and frequency content, which makes the choice of augmentation substantially more delicate than in standard photographic datasets.

Data augmentation has become a central component of modern deep learning because it can enlarge the effective support of the training distribution and improve generalization under limited data \cite{shorten2019survey,cubuk2019autoaugment}. Most widely used policies, however, were developed for natural images and encode assumptions of label invariance to pose changes, photometric perturbations, occlusion, or synthetic corruption. The robustness literature has repeatedly shown that these gains are not uniform across corruption types and do not necessarily transfer to natural distribution shifts \cite{hendrycks2019benchmarking,rusak2020simple,taori2020measuring}. Frequency-domain analyses further indicate that augmentation can alter the spectral sensitivity of convolutional models in nontrivial ways \cite{yin2019fourier}. These observations are especially relevant in coherent imaging, where a transformation that appears visually mild may nevertheless alter the physics-bearing statistics of the measurement.

Recent work in scientific imaging has reinforced the need for domain-specific robustness analysis. Deep learning is now widely used in holography and coherent imaging \cite{rivenson2019deep}, and robustness studies in microscopy have shown that conclusions derived from natural-image benchmarks do not automatically transfer to scientific acquisition settings \cite{zhong2024benchmarking}. Yet, despite growing interest in speckle-based learning, there remains limited understanding of which augmentation families preserve the information required for accurate classification and which ones distort that information beyond usefulness. The central question in this paper is therefore not simply whether augmentation helps, but which physical properties of speckle must be preserved for augmentation to remain beneficial.

\begin{figure}[t]
\centering
\includegraphics[width=\columnwidth]{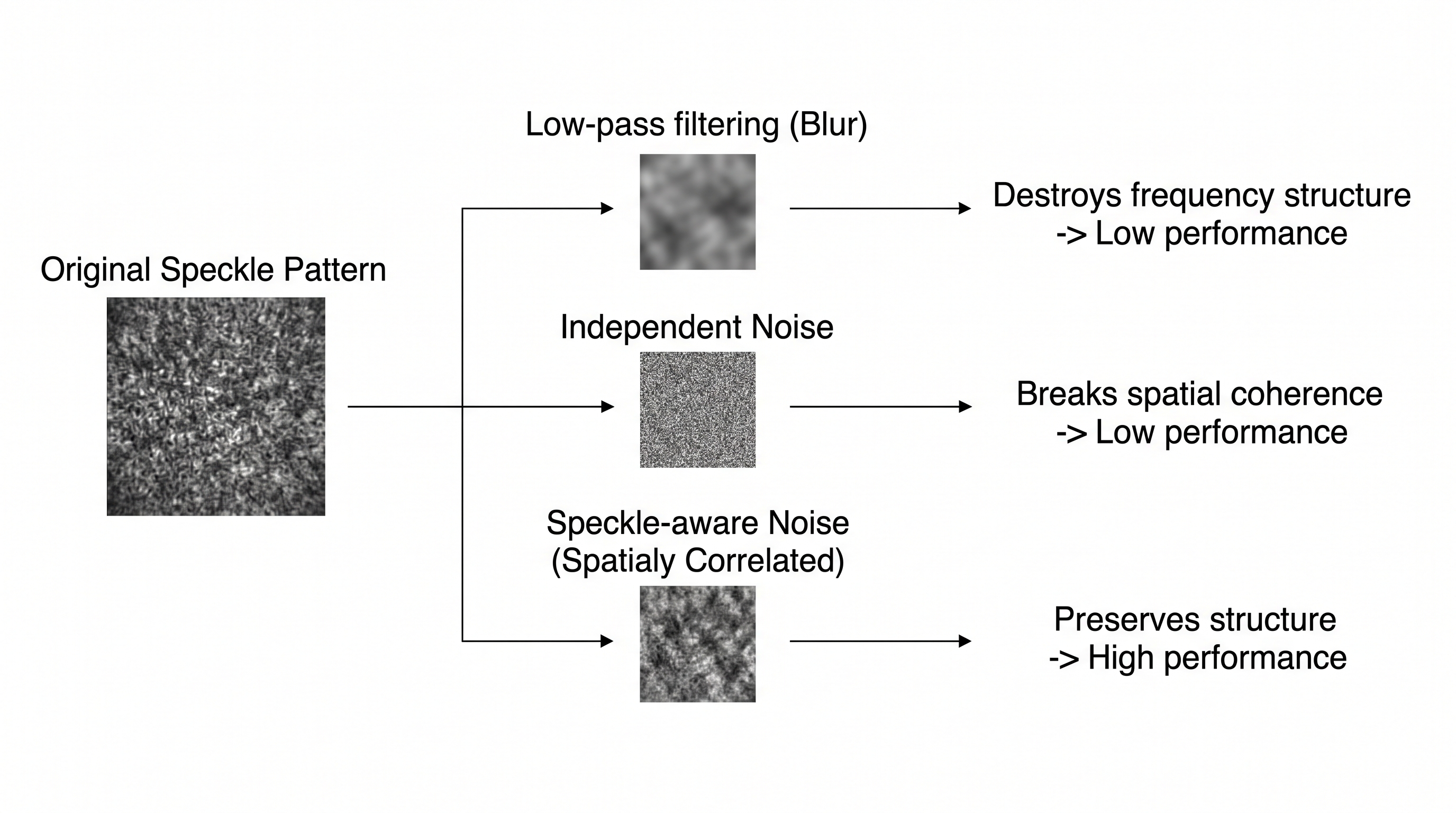}
\caption{Conceptual summary of the structural-preservation hypothesis examined in this work. Low-pass blur suppresses frequency content, independent pixel-wise noise breaks local coherence, and speckle-aware perturbations preserve spatial organization while introducing controlled variability.}
\label{fig:concept}
\end{figure}

The present study addresses this question through a controlled, parametric analysis of augmentation effects in laser speckle material classification. Using the SensiCut dataset and two representative convolutional backbones, ResNet18 and EfficientNet-B0, we vary rotation, blur, independent noise, spatially correlated speckle-aware noise, intensity jitter, and spatial masking over continuous ranges and fit architecture-specific regression models to relate augmentation parameters to macro F1-score. This design allows augmentation to be studied as a structured perturbation problem rather than as a set of isolated recipes. The resulting manuscript argues, and the experiments support, that augmentation effectiveness in speckle imaging is governed primarily by structural preservation of spatial coherence and frequency content. Figure~\ref{fig:concept} summarizes this working hypothesis and the three main transformation behaviors evaluated in the study.

\section{Related Work}

The literature on augmentation for deep learning is extensive, but much of it is grounded in natural-image benchmarks. Surveys such as that of Shorten and Khoshgoftaar \cite{shorten2019survey} document the broad utility of geometric and photometric augmentation, while policy-search methods such as AutoAugment show that learned augmentation schedules can improve benchmark accuracy when the search space is well aligned with the task \cite{cubuk2019autoaugment}. Occlusion-style perturbations, including random erasing, have likewise been effective as regularizers when partial spatial removal is a plausible nuisance factor \cite{zhong2020random}. The common thread across these methods is an implicit assumption that semantic identity is preserved under the augmentation family being applied.

Robustness studies complicate this picture by showing that augmentation gains are strongly corruption-dependent. Hendrycks and Dietterich demonstrated that many high-performing image classifiers remain fragile to blur, noise, and other common corruptions \cite{hendrycks2019benchmarking}. Rusak \emph{et al.} showed that robustness can improve when training perturbations more closely match the corruption family encountered at test time \cite{rusak2020simple}. Taori \emph{et al.} further reported that robustness to synthetic corruptions and robustness to natural distribution shifts are related but distinct objectives \cite{taori2020measuring}. From a spectral perspective, Yin \emph{et al.} showed that corruption sensitivity can be understood in part through the Fourier characteristics of the perturbation and the model \cite{yin2019fourier}. Taken together, these studies imply that augmentation should be interpreted as a distribution-design problem rather than a universally beneficial preprocessing step.

The coherent imaging and speckle literature provides additional motivation for task-specific analysis. Reviews of laser speckle contrast imaging emphasize that speckle statistics are tightly coupled to optical coherence, acquisition geometry, and sample dynamics \cite{boas2010laser,briers2013laser}. At the same time, deep learning is increasingly used in coherent imaging pipelines, where the interaction between network priors and wave-based measurements has become an active research topic \cite{rivenson2019deep}. Learning from speckle patterns has already proven useful in several applications, including material-aware laser cutting \cite{salem2023material, dogan2021sensicut, salem2025safer} and the interpretation of coherent X-ray speckle patterns \cite{shen2023machine}. Subsequent application-specific studies have extended this line of work to laser-cutting material classification with red-laser speckle acquisition, hazardous-material detection during laser cutting, robustness to laser-color changes, and lightweight edge-compatible CNN design for speckle-based recognition \cite{salem2023material,salem2023hazardous,salem2025edge}. However, the augmentation question remains underdeveloped. Existing work demonstrates that speckle can be informative, but it does not systematically identify which augmentation mechanisms preserve the statistics that make speckle informative in the first place. This gap motivates the present study.

\begin{figure*}[t]
\centering
\includegraphics[width=\textwidth]{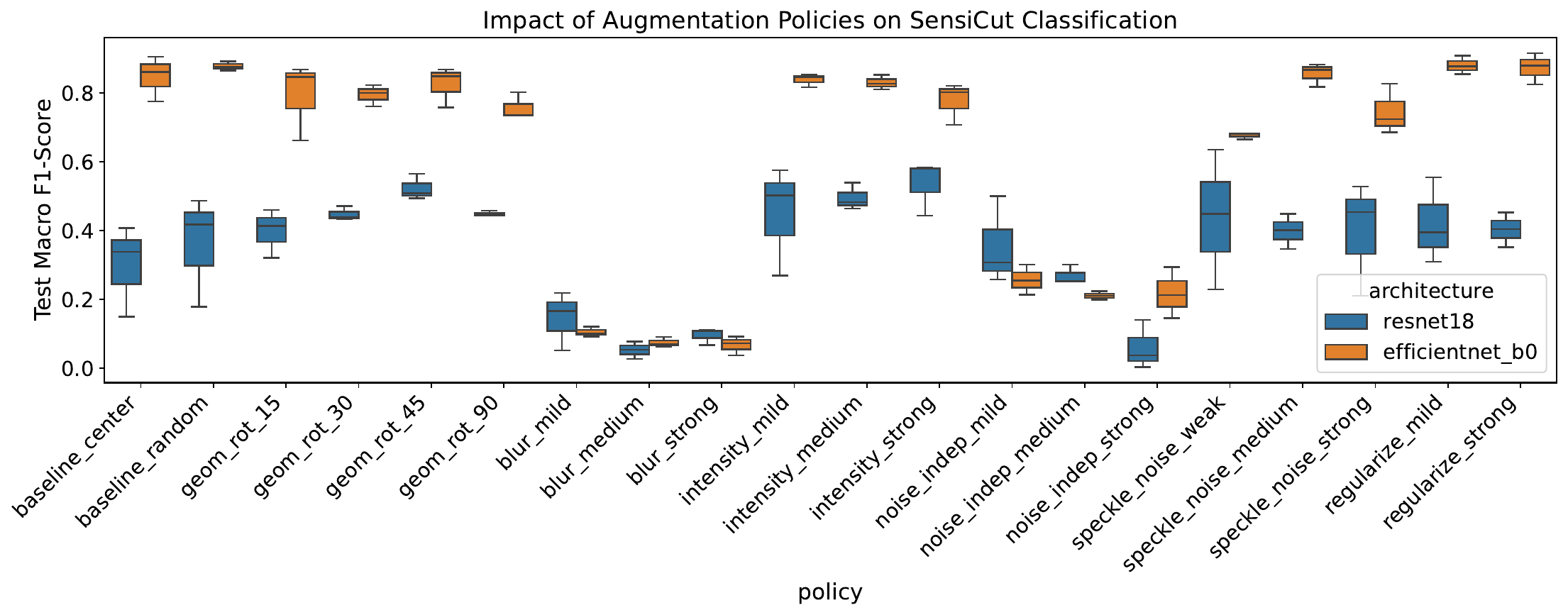}
\caption{Distribution of test macro F1 across augmentation policies and random seeds. The main qualitative pattern is the collapse caused by blur and independent noise, contrasted with the more favorable behavior of moderate structure-preserving policies.}
\label{fig:boxplot}
\end{figure*}

\section{Methodology}

The experiments were designed to isolate how different augmentation mechanisms alter performance in speckle-based material classification.

\subsection{Dataset Preparation}

We used the public SensiCut dataset introduced by Dogan \emph{et al.} \cite{dogan2021sensicut}. The dataset consists of laser speckle images acquired from multiple physical material surfaces using a controlled optical sensing system. The SensiCut taxonomy naturally supports two complementary label granularities. In the fine-grained formulation, each physical sample is treated as an independent class, resulting in 59 sample-level classes that capture intra-material variability associated with surface finish, manufacturing conditions, and acquisition instances. In the coarse formulation, the same samples are grouped according to their underlying material type, yielding 30 classes consistent with the original material taxonomy. For the augmentation analysis reported in this paper, results are presented for the 30-class material-level formulation, while the 59-sample hierarchy is retained during dataset preparation to clarify label granularity and split semantics.

After curation for the present study, the working dataset comprised 39{,}609 images. We created training, validation, and test subsets using a stratified image-level split of 70\%, 15\%, and 15\%, respectively, so that every class remained represented while image instances stayed disjoint across subsets. The resulting evaluation therefore measures generalization to previously unseen observations of the same physical samples rather than to entirely unseen material instances. Class frequencies were approximately balanced in both label formulations, with only minor deviations attributable to acquisition constraints. Performance was therefore evaluated using macro-averaged F1-score.

Preprocessing was designed to preserve the structural characteristics of speckle patterns. Interpolative resizing was avoided because it can introduce smoothing and aliasing that alter the frequency content of the signal. Instead, fixed-size $224 \times 224$ patches were extracted from the native images by cropping. Within the experimental design, center and random crops were retained as explicit baselines, allowing positional variability to be introduced without disturbing the native spatial statistics of the speckle field. This formulation supports controlled evaluation of augmentation robustness while minimizing confounding effects arising from sample-level distribution shifts.

\subsection{Model Architectures}

To test whether the augmentation effects were architecture-specific or instead reflected broader properties of the data, we used two convolutional architectures. ResNet18 was chosen as a canonical residual network with comparatively modest depth and a well-established optimization profile \cite{he2016deep}. EfficientNet-B0 was used as a stronger baseline that combines compound scaling with a more parameter-efficient architecture \cite{tan2019efficientnet}. Evaluating both models allows the analysis to compare a conventional residual backbone with a more modern architecture that often exhibits stronger baseline accuracy.

\subsection{Parametric Augmentation and Training}

Augmentation was formulated parametrically so that each transformation could be interpreted as a controllable perturbation of the input distribution rather than as a binary on/off choice. The geometric variable was in-plane rotation, parameterized by angle $\theta$. Frequency-domain distortion was introduced with Gaussian blur of standard deviation $\sigma_b$, which acts as a low-pass filter. Independent stochastic perturbation was implemented as additive Gaussian noise with standard deviation $\sigma_n$, sampled independently at each pixel. Speckle-aware perturbation was implemented as spatially correlated noise with correlation scale $\sigma_s$, constructed to introduce local variation while preserving neighborhood structure. Intensity jitter of magnitude $j$ modulated global brightness and contrast, and spatial masking was controlled by erase probability $p_e$, following the general idea of random erasing \cite{zhong2020random}. In addition to these augmented policies, center-crop and random-crop baselines were retained for comparison. The full design comprised 20 policies, evaluated for both architectures across three random seeds, for a total of 120 trained models.

All models were trained under the same optimization protocol, with identical stopping criteria and seed averaging so that comparisons reflect augmentation effects rather than changes in optimization procedure. Performance was measured on the held-out test set using macro F1-score, which weights all classes equally and is therefore appropriate when per-class behavior is as important as aggregate accuracy. For each policy, the reported score corresponds to the mean across three seeds.

\subsection{Regression Formulation}

To quantify the contribution of each augmentation variable, we fit a separate ordinary least squares model for each architecture. The response variable was test macro F1-score, and the predictors were the continuous augmentation parameters:

\begin{equation}
\label{eq:ols}
y = \beta_0 + \beta_1\theta + \beta_2\sigma_b + \beta_3\sigma_n + \beta_4\sigma_s + \beta_5 p_e + \beta_6 j + \varepsilon,
\end{equation}

where $y$ denotes mean test macro F1-score and $\varepsilon$ is the residual term. This regression formulation is useful because it estimates the direction and strength of each transformation while controlling for the others. It also makes it possible to distinguish whether a policy performs well because of one dominant perturbation or because several transformations act synergistically.

\begin{figure*}[t]
\centering
\subfloat[FFT distance\label{fig:fft-corr}]{\includegraphics[width=0.32\textwidth]{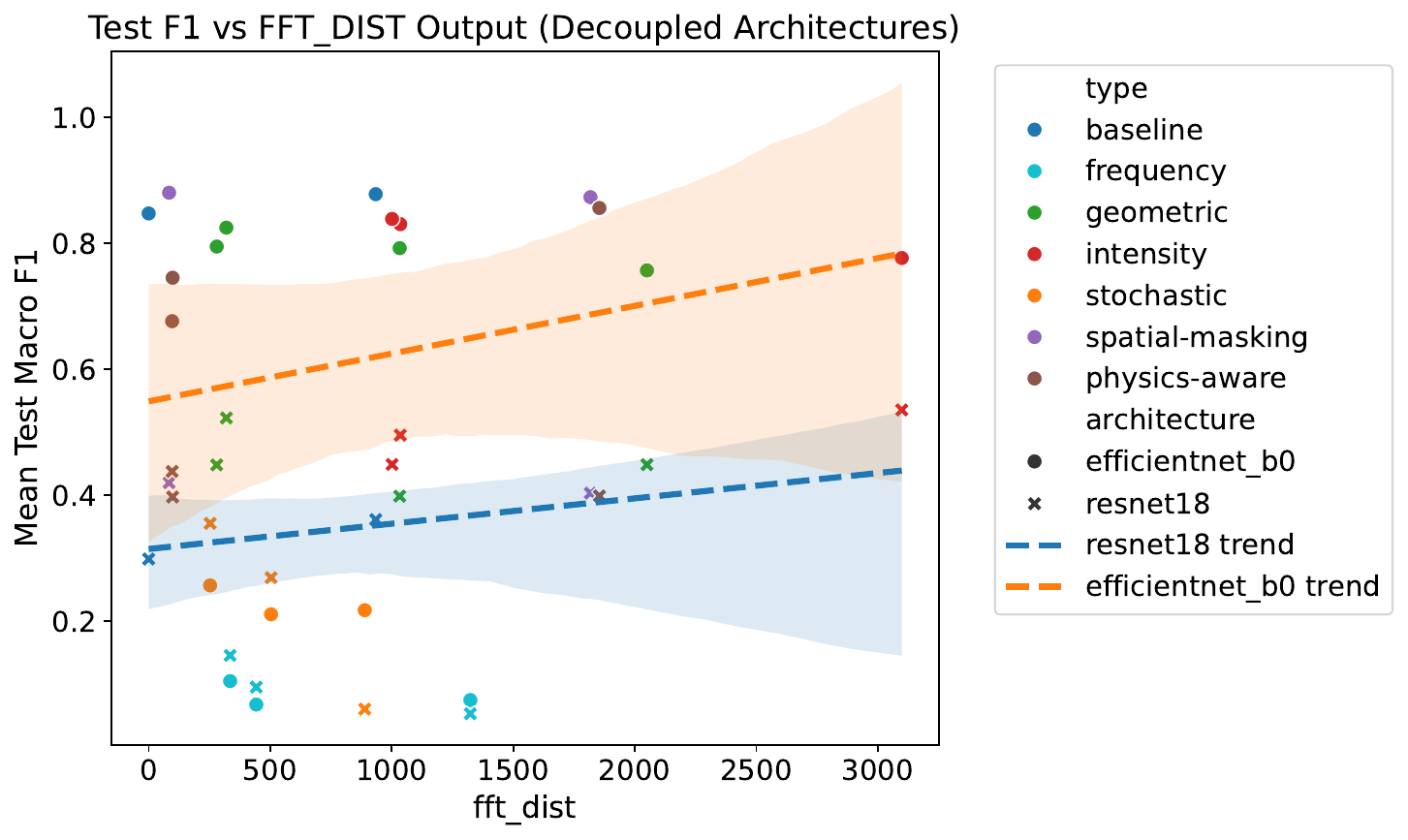}}\hfil
\subfloat[Histogram distance\label{fig:hist-corr}]{\includegraphics[width=0.32\textwidth]{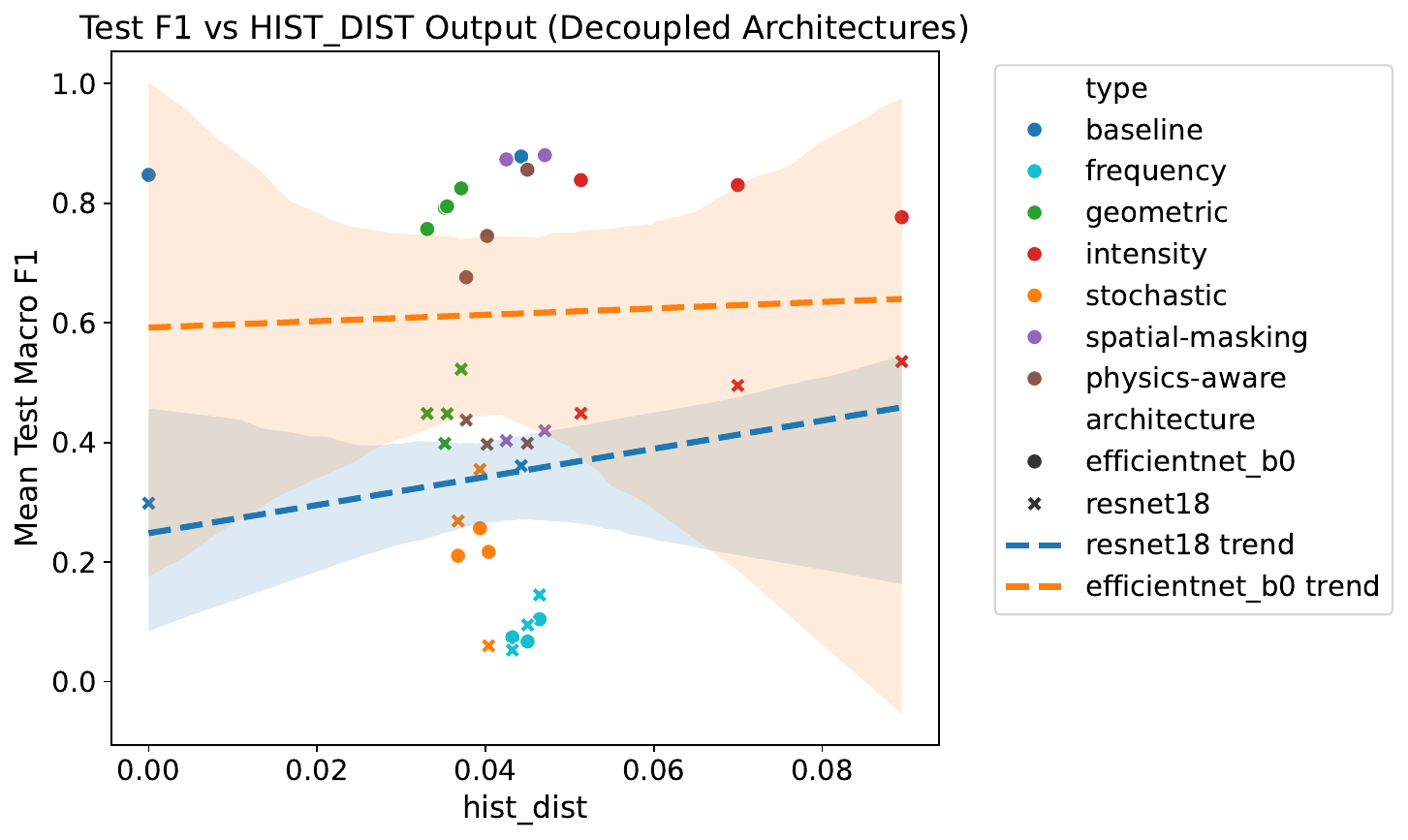}}\hfil
\subfloat[SSIM\label{fig:ssim-corr}]{\includegraphics[width=0.32\textwidth]{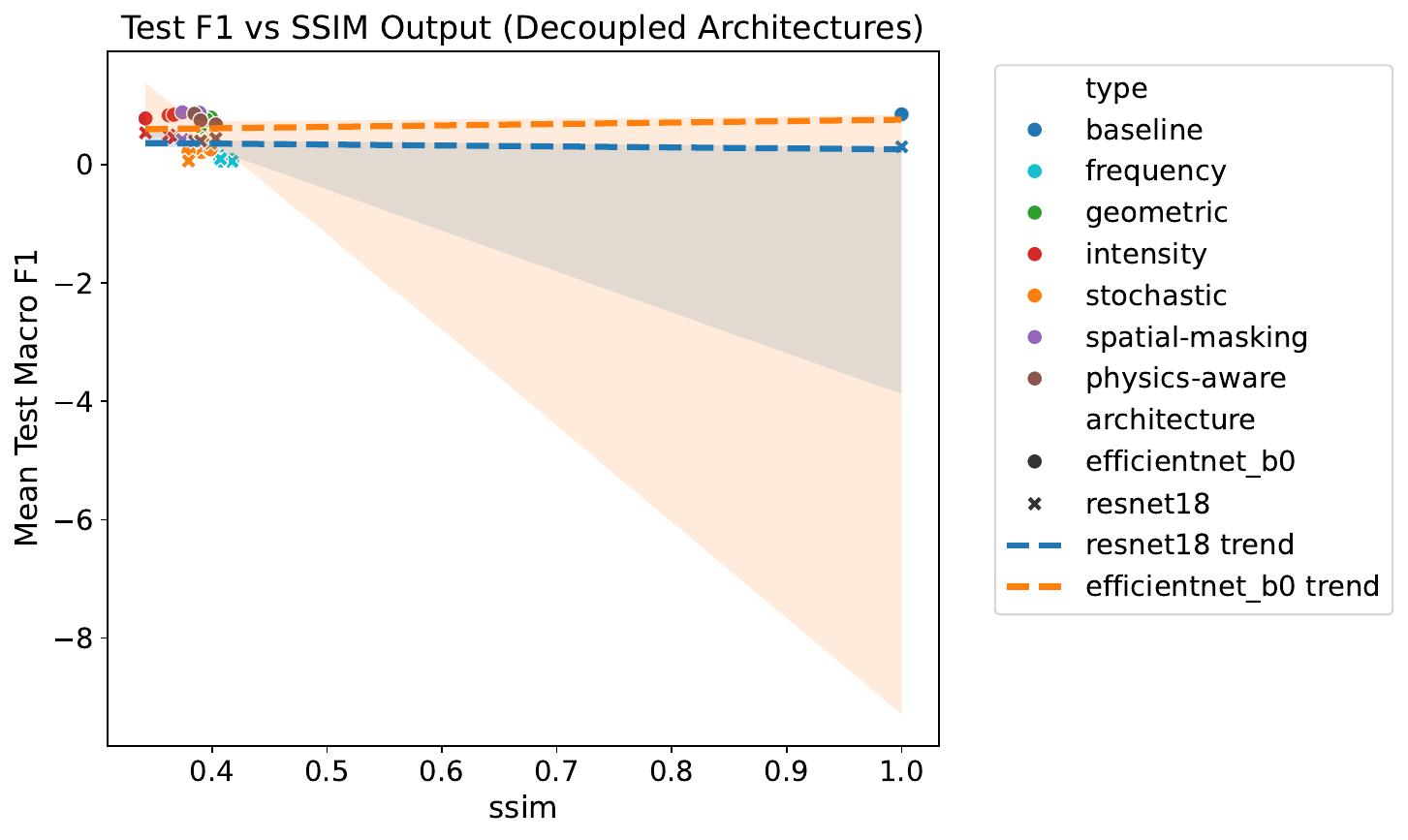}}
\caption{Relationships between test macro F1 and scalar descriptors of augmentation-induced distortion. The plots show that structural descriptors are more informative than histogram change alone, while also indicating that no single scalar metric fully captures the interaction between augmentation and speckle statistics.}
\label{fig:distortion-correlations}
\end{figure*}

\section{Results}

\begin{figure}[t]
\centering
\includegraphics[width=\columnwidth]{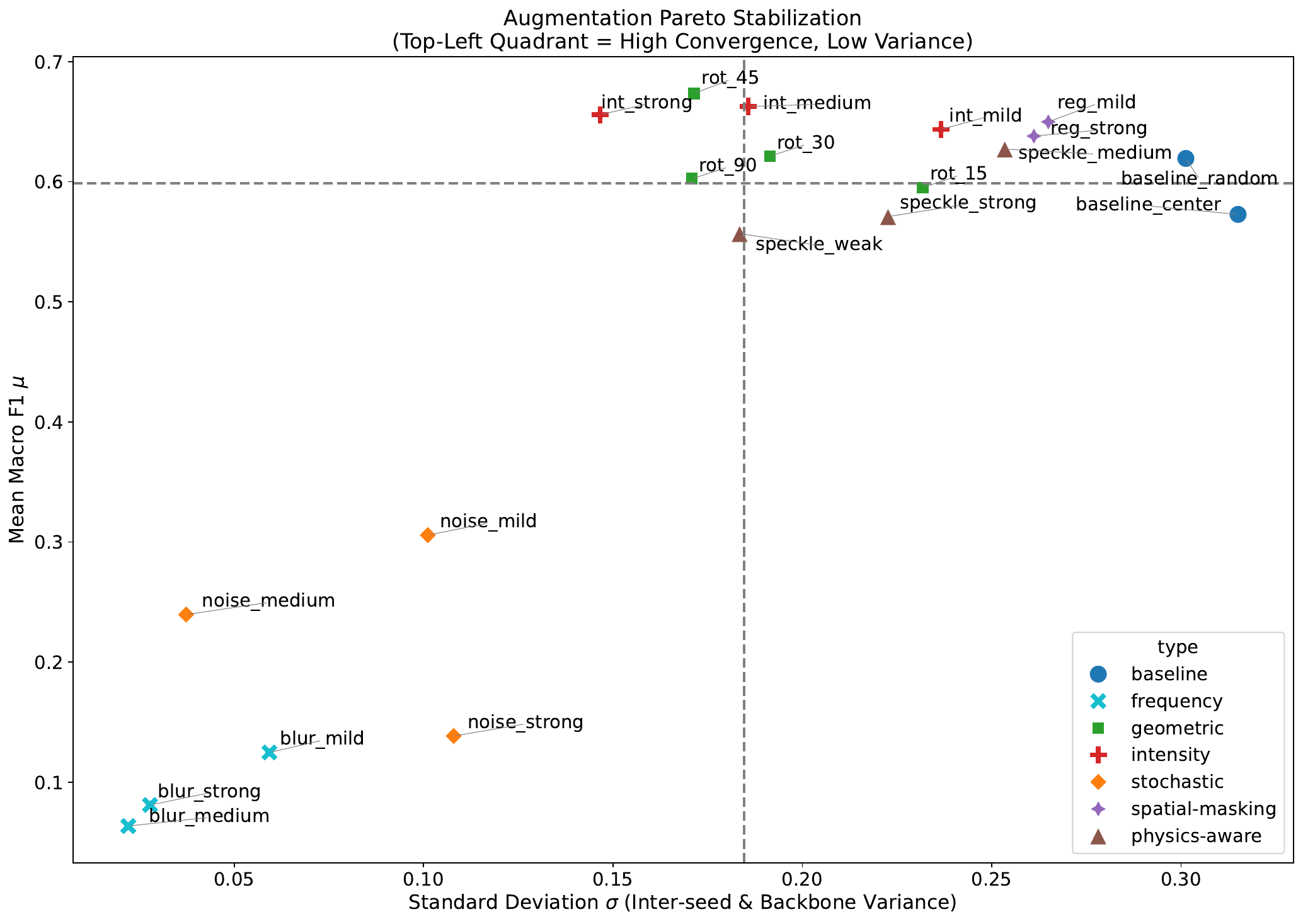}
\caption{Pareto view of mean test macro F1 versus variance across augmentation policies. Policies near the favorable frontier combine strong average performance with stable behavior across seeds.}
\label{fig:pareto}
\end{figure}

The augmentation landscape is summarized qualitatively by the performance distribution in Fig.~\ref{fig:boxplot} and the mean--variance view in Fig.~\ref{fig:pareto}. EfficientNet-B0 achieved higher absolute macro F1 than ResNet18 under most policies, but the ranking of harmful and beneficial perturbations was similar across both architectures. Gaussian blur consistently occupied the lowest-performing region, with mean macro F1 values below 0.11 for all three blur settings in both models. Independent Gaussian noise also substantially degraded performance, especially for EfficientNet-B0, for which all independent-noise policies remained near 0.23 macro F1 or below. By contrast, moderate structure-preserving policies remained competitive. For EfficientNet-B0, moderate spatial masking and medium speckle-aware noise were among the strongest configurations, alongside the random-crop baseline. For ResNet18, performance was more variable in absolute terms, but blur and independent noise again defined the low-performance regime, while moderate rotations and intensity modulation were comparatively more tolerant.

The regression analysis in Table~\ref{tab:regression} formalizes these observations. For EfficientNet-B0, the model explained 79.6\% of the observed variance in macro F1 ($R^2=0.796$, model $p=7.10\times 10^{-4}$). Gaussian blur had a strongly negative coefficient ($\beta=-0.2693$, $p<0.001$), indicating that low-pass filtering systematically reduced performance. Independent Gaussian noise was also significantly harmful ($\beta=-3.2102$, $p=0.003$). In contrast, the speckle-aware perturbation term was positive and significant ($\beta=0.2233$, $p=0.004$), showing that spatially correlated variability improved robustness when it preserved local structure. Rotation, erase probability, and intensity jitter were not significant in the EfficientNet-B0 model.

For ResNet18, the fitted model explained 87.9\% of the variance in test macro F1 ($R^2=0.879$, model $p=2.82\times 10^{-5}$). The signs of the key coefficients matched those of EfficientNet-B0. Blur again showed a strong negative effect ($\beta=-0.1217$, $p<0.001$), and independent noise remained significantly harmful ($\beta=-1.4016$, $p=0.001$). The coefficient of the speckle-aware term was positive and statistically significant ($\beta=0.0984$, $p=0.001$). Rotation was marginal but did not cross the 0.05 threshold, and spatial masking remained nonsignificant after controlling for the other variables. Intensity jitter differed from EfficientNet-B0 in that it showed a positive association with performance in ResNet18 ($\beta=0.2210$, $p=0.014$), suggesting a model-dependent tolerance to global photometric modulation. Nevertheless, the dominant pattern across architectures was the same: transformations that suppressed frequency structure or broke local coherence were detrimental, whereas structure-preserving perturbations were beneficial.

The distortion-oriented plots in Fig.~\ref{fig:distortion-correlations} and the Pareto analysis in Fig.~\ref{fig:pareto} provide a complementary view. Among the scalar distortion descriptors, the SSIM-based view showed the clearest monotonic association with macro F1 across both architectures, whereas histogram distance was comparatively diffuse, indicating that simple intensity-distribution changes alone do not explain the performance ordering. The Fourier-distance view is consistent with the regression result that frequency-disruptive transformations are harmful, but it also shows that spectral change by itself is not sufficient to predict performance without considering spatial organization. The Pareto plot further shows that the most useful policies are not merely those that maximize average F1, but those that do so without inflating variance across seeds. In this regard, moderate structure-preserving perturbations occupy a more favorable frontier than blur or independent noise, which are simultaneously low-performing and unstable.

\begin{table*}[t]
\centering
\caption{Ordinary least squares coefficients relating augmentation parameters to test macro F1-score. Separate models were fitted for EfficientNet-B0 and ResNet18.}
\label{tab:regression}
\renewcommand{\arraystretch}{1.1}
\setlength{\tabcolsep}{10pt}
\begin{tabular}{lcccc}
\toprule
Predictor & EfficientNet-B0 coef. & EfficientNet-B0 $p$ & ResNet18 coef. & ResNet18 $p$ \\
\midrule
Rotation $\theta$ & 0.0013 & 0.536 & 0.0014 & 0.070 \\
Blur $\sigma_b$ & -0.2693 & $<0.001$ & -0.1217 & $<0.001$ \\
Independent noise $\sigma_n$ & -3.2102 & 0.003 & -1.4016 & 0.001 \\
Speckle-aware noise $\sigma_s$ & 0.2233 & 0.004 & 0.0984 & 0.001 \\
Erase probability $p_e$ & 0.6830 & 0.287 & 0.1351 & 0.558 \\
Intensity jitter $j$ & 0.1589 & 0.468 & 0.2210 & 0.014 \\
\midrule
$R^2$ & 0.796 & --- & 0.879 & --- \\
Model $p$-value & $7.10\times 10^{-4}$ & --- & $2.82\times 10^{-5}$ & --- \\
\bottomrule
\end{tabular}
\end{table*}

\section{Discussion}

The results support a simple but consequential interpretation: in laser speckle classification, the utility of augmentation depends less on perturbation magnitude than on whether the perturbation respects the statistical structure generated by coherent scattering. Gaussian blur was consistently harmful because it removes high-frequency content and reduces local contrast fluctuations that are intrinsic to speckle formation. In natural images, blur may remove nuisance texture while leaving semantic shape relatively intact. In speckle imagery, by contrast, the fine-scale fluctuations are not nuisance details but part of the class-discriminative signal itself \cite{boas2010laser,briers2013laser}. The negative blur coefficients observed in both architectures are therefore physically interpretable rather than merely empirical.

Independent Gaussian noise failed for a related but distinct reason. Even when its variance is controlled, pixel-wise noise perturbs each location without regard to the correlation structure that coherent interference induces across neighboring pixels. This creates samples that are statistically implausible under the acquisition process. The fact that independent noise was significantly harmful in both regression models suggests that robustness in this domain cannot be improved by arbitrary stochasticity. What matters is not simply exposing the model to variation, but exposing it to variation that lies on a plausible manifold for speckle formation. This conclusion is consistent with broader robustness findings showing that mismatch between training perturbations and evaluation distortions can negate the intended regularization benefit \cite{rusak2020simple,taori2020measuring}.

The positive effect of speckle-aware perturbations is therefore particularly informative. Spatially correlated noise introduces variability while preserving local organization, and in doing so it acts more like a physically plausible perturbation than like an artificial corruption. The effect was significant in both EfficientNet-B0 and ResNet18, despite the substantial difference in their absolute performance levels. This sign consistency indicates that the phenomenon is data-centric rather than architecture-specific. It also suggests that physics-aware augmentation need not be highly elaborate to be effective; even relatively simple correlation-preserving perturbations can move the training distribution in a beneficial direction when they respect the structure of the sensing process.

The remaining augmentation families help refine this interpretation. Rotation produced small coefficients and no clear regression significance, which suggests that modest geometric variation is neither especially beneficial nor especially destructive once the stronger structural effects are accounted for. Spatial masking performed well in some settings, particularly for EfficientNet-B0, but its regression coefficient was not significant. This indicates that masking may function as a secondary regularizer rather than as a primary mechanism of robustness in this domain. The positive jitter coefficient for ResNet18, absent in EfficientNet-B0, points to a model-dependent tolerance to global photometric shifts. Importantly, however, these secondary effects do not overturn the central result. Across both architectures, the dominant determinants of augmentation success are preservation of spatial coherence and retention of informative frequency structure.

These findings have practical implications for coherent optical sensing. Standard augmentation search spaces imported from natural-image classification are unlikely to be optimal when the measurement itself is a stochastic interference pattern. Instead, augmentation design should be constrained by optical plausibility. For speckle-based recognition, this means favoring transformations that preserve correlation length, local continuity, and spectral envelope while avoiding operations that sever those relationships. More generally, the results align with emerging evidence from coherent and scientific imaging that model robustness depends on respecting the generative physics of the measurement process \cite{rivenson2019deep,zhong2024benchmarking,shen2023machine}. Future work should therefore move beyond generic policy transfer toward adaptive augmentation schemes informed by wave-optics models, class-conditional distortion tolerances, or learned measures of speckle fidelity.

\section{Conclusion}

This paper presented a structured analysis of data augmentation for laser speckle material classification using the SensiCut dataset and two convolutional architectures. By treating augmentation parameters as continuous variables and relating them to macro F1-score through ordinary least squares regression, the study moved beyond recipe-level comparison and identified the dominant factors associated with performance changes. The results were consistent across EfficientNet-B0 and ResNet18: Gaussian blur and independent pixel-wise noise degraded classification because they suppress frequency information or disrupt local coherence, whereas spatially correlated speckle-aware perturbations improved robustness by preserving the organization of the speckle field.

The main implication is that augmentation effectiveness in coherent imaging is governed by structural preservation. In speckle-based learning, useful variability is not arbitrary variability. It must remain compatible with the spatial and frequency statistics induced by coherent scattering. This principle provides a concrete foundation for future augmentation design in optical sensing, including physics-aware policies, adaptive strategies tied to measured distortion, and joint learning frameworks that incorporate forward models of image formation.

\bibliographystyle{IEEEtran}
\bibliography{references}

@article{shorten2019survey,
  author = {Shorten, Connor and Khoshgoftaar, Taghi M.},
  title = {A Survey on Image Data Augmentation for Deep Learning},
  journal = {Journal of Big Data},
  volume = {6},
  number = {1},
  pages = {60},
  year = {2019},
  doi = {10.1186/s40537-019-0197-0}
}

@inproceedings{cubuk2019autoaugment,
  author = {Cubuk, Ekin D. and Zoph, Barret and Mane, Dandelion and Vasudevan, Vijay and Le, Quoc V.},
  title = {AutoAugment: Learning Augmentation Strategies From Data},
  booktitle = {Proceedings of the IEEE/CVF Conference on Computer Vision and Pattern Recognition},
  pages = {113--123},
  year = {2019},
  doi = {10.1109/CVPR.2019.00020}
}

@inproceedings{zhong2020random,
  author = {Zhong, Zhun and Zheng, Liang and Kang, Guoliang and Li, Shaozi and Yang, Yi},
  title = {Random Erasing Data Augmentation},
  booktitle = {Proceedings of the AAAI Conference on Artificial Intelligence},
  volume = {34},
  number = {7},
  pages = {13001--13008},
  year = {2020},
  doi = {10.1609/aaai.v34i07.7000}
}

@inproceedings{hendrycks2019benchmarking,
  author = {Hendrycks, Dan and Dietterich, Thomas},
  title = {Benchmarking Neural Network Robustness to Common Corruptions and Perturbations},
  booktitle = {International Conference on Learning Representations},
  year = {2019}
}

@incollection{rusak2020simple,
  author = {Rusak, Evgenia and Schott, Lukas and Zimmermann, Roland S. and Bitterwolf, Julian and Bringmann, Oliver and Bethge, Matthias and Brendel, Wieland},
  title = {A Simple Way to Make Neural Networks Robust Against Diverse Image Corruptions},
  booktitle = {Computer Vision -- ECCV 2020},
  series = {Lecture Notes in Computer Science},
  volume = {12348},
  pages = {53--69},
  year = {2020},
  publisher = {Springer},
  doi = {10.1007/978-3-030-58580-8_4}
}

@inproceedings{taori2020measuring,
  author = {Taori, Rohan and Dave, Achal and Shankar, Vaishaal and Carlini, Nicholas and Recht, Benjamin and Schmidt, Ludwig},
  title = {Measuring Robustness to Natural Distribution Shifts in Image Classification},
  booktitle = {Advances in Neural Information Processing Systems},
  volume = {33},
  year = {2020}
}

@inproceedings{yin2019fourier,
  author = {Yin, Dong and Lopes, Raphael Gontijo and Shlens, Jon and Cubuk, Ekin Dogus and Gilmer, Justin},
  title = {A Fourier Perspective on Model Robustness in Computer Vision},
  booktitle = {Advances in Neural Information Processing Systems},
  volume = {32},
  pages = {13255--13265},
  year = {2019}
}

@article{boas2010laser,
  author = {Boas, David A. and Dunn, Andrew K.},
  title = {Laser Speckle Contrast Imaging in Biomedical Optics},
  journal = {Journal of Biomedical Optics},
  volume = {15},
  number = {1},
  pages = {011109},
  year = {2010},
  doi = {10.1117/1.3285504}
}

@article{briers2013laser,
  author = {Briers, David and Duncan, Donald D. and Hirst, Evan and Kirkpatrick, Sean J. and Larsson, Marcus and Steenbergen, Wiendelt and St\"romberg, Tomas and Thompson, Oliver B.},
  title = {Laser Speckle Contrast Imaging: Theoretical and Practical Limitations},
  journal = {Journal of Biomedical Optics},
  volume = {18},
  number = {6},
  pages = {066018},
  year = {2013},
  doi = {10.1117/1.JBO.18.6.066018}
}

@article{rivenson2019deep,
  author = {Rivenson, Yair and Wu, Yichen and Ozcan, Aydogan},
  title = {Deep Learning in Holography and Coherent Imaging},
  journal = {Light: Science \& Applications},
  volume = {8},
  pages = {85},
  year = {2019},
  doi = {10.1038/s41377-019-0196-0}
}

@inproceedings{dogan2021sensicut,
  author = {Dogan, Mustafa Doga and Acevedo Colon, Steven Vidal and Sinha, Varnika and Ak{\c{s}}it, Kaan and Mueller, Stefanie},
  title = {SensiCut: Material-Aware Laser Cutting Using Speckle Sensing and Deep Learning},
  booktitle = {The 34th Annual ACM Symposium on User Interface Software and Technology},
  pages = {24--38},
  year = {2021},
  doi = {10.1145/3472749.3474733}
}

@inproceedings{salem2023material,
  author = {Salem, Mohamed Abdallah and Elshenawy, Ahmed and Ashour, Hamdy Ahmed},
  title = {Material Classification in Laser Cutting Using Deep Learning},
  booktitle = {2023 Intelligent Methods, Systems, and Applications (IMSA)},
  pages = {167--173},
  year = {2023},
  organization = {IEEE}
}

@article{salem2023hazardous,
  author = {Salem, Mohamed Abdallah and ElShenawy, Ahmed K. and Ashour, Hamdy A.},
  title = {Detection of Hazardous Materials in Laser Cutting Using Deep Learning and Speckle Sensing},
  journal = {The International Archives of the Photogrammetry, Remote Sensing and Spatial Information Sciences},
  volume = {XLVIII-1/W2-2023},
  pages = {497--503},
  year = {2023},
  doi = {10.5194/isprs-archives-XLVIII-1-W2-2023-497-2023}
}

@INPROCEEDINGS{salem2025edge,
  author={Salem, Mohamed Abdallah and Diab, Nourhan Zein},
  booktitle={2025 35th International Conference on Computer Theory and Applications (ICCTA)}, 
  title={Efficient Edge-Compatible CNN for Speckle-Based Material Recognition in Laser Cutting Systems}, 
  year={2025},
  volume={},
  number={},
  pages={296-301},
  doi={10.1109/ICCTA68914.2025.11519797}}

@misc{salem2025safer,
      title={Towards a Safer and Sustainable Manufacturing Process: Material classification in Laser Cutting Using Deep Learning}, 
      author={Mohamed Abdallah Salem and Hamdy Ahmed Ashur and Ahmed Elshinnawy},
      year={2025},
      eprint={2511.16026},
      archivePrefix={arXiv},
      primaryClass={cs.CV},
      url={https://arxiv.org/abs/2511.16026}, 
}

@article{zhong2024benchmarking,
  author = {Zhong, Liqun and Li, Lingrui and Yang, Ge},
  title = {Benchmarking Robustness of Deep Neural Networks in Semantic Segmentation of Fluorescence Microscopy Images},
  journal = {BMC Bioinformatics},
  volume = {25},
  number = {1},
  pages = {269},
  year = {2024},
  doi = {10.1186/s12859-024-05894-4}
}

@article{shen2023machine,
  author = {Shen, Mingren and Sheyfer, Dina and Loeffler, Troy D. and Sankaranarayanan, Subramanian K. R. S. and Stephenson, G. Brian and Chan, Maria K. Y. and Morgan, Dane},
  title = {Machine Learning for Interpreting Coherent X-Ray Speckle Patterns},
  journal = {Computational Materials Science},
  volume = {230},
  pages = {112500},
  year = {2023},
  doi = {10.1016/j.commatsci.2023.112500}
}

@inproceedings{he2016deep,
  author = {He, Kaiming and Zhang, Xiangyu and Ren, Shaoqing and Sun, Jian},
  title = {Deep Residual Learning for Image Recognition},
  booktitle = {Proceedings of the IEEE Conference on Computer Vision and Pattern Recognition},
  pages = {770--778},
  year = {2016}
}

@inproceedings{tan2019efficientnet,
  author = {Tan, Mingxing and Le, Quoc},
  title = {EfficientNet: Rethinking Model Scaling for Convolutional Neural Networks},
  booktitle = {Proceedings of the 36th International Conference on Machine Learning},
  series = {Proceedings of Machine Learning Research},
  volume = {97},
  pages = {6105--6114},
  year = {2019},
  publisher = {PMLR}
}

\end{document}